\documentclass[letterpaper]{article}

\usepackage{natbib,alifeconf}  %
\usepackage{amsmath}
\usepackage{url,hyperref,cleveref}
\usepackage{booktabs}

\title{Neural Cellular Automata for ARC-AGI}

\author{
    Kevin Xu$^{1}$,
    Risto Miikkulainen$^{1}$ \\
    \mbox{}\\
    $^1$The University of Texas at Austin, Austin, TX, USA \\
    \texttt{kx@utexas.edu, risto@cs.utexas.edu}
}

\begin{document}

\maketitle

\begin{abstract}
Cellular automata and their differentiable counterparts, Neural Cellular Automata (NCA), are highly expressive and capable of surprisingly complex behaviors. This paper explores how NCAs perform when applied to tasks requiring precise transformations and few-shot generalization, using the Abstraction and Reasoning Corpus for Artificial General Intelligence (ARC-AGI) as a domain that challenges their capabilities in ways not previously explored. Specifically, this paper uses gradient-based training to learn iterative update rules that transform input grids into their outputs from the training examples and then applies them to the test inputs. Results suggest that gradient-trained NCA models are a promising and efficient approach to a range of abstract grid-based tasks from ARC. Along with discussing the impacts of various design modifications and training constraints, this work examines the behavior and properties of NCAs applied to ARC to give insights for broader applications of self-organizing systems.
\end{abstract}

\section{Introduction}
Most modern machine learning systems rely on large, centrally organized models trained with vast amounts of data. In contrast, many of the traits associated with intelligent behavior, such as robustness, adaptability, and generalization from few examples, appear in nature as emergent outcomes of self-organizing local interactions. This raises a question: without global control or symbolic scaffolding, can self-organizing neural models serve as viable substrates for solving structured tasks?

\begin{figure}[!t]
  \centering
  \includegraphics[width=\linewidth]{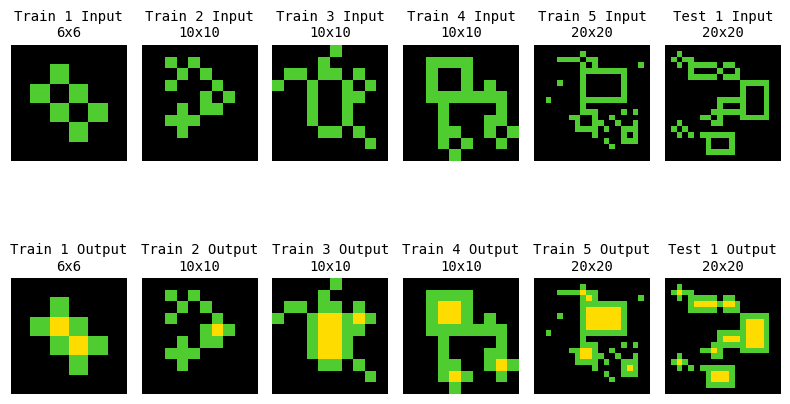}
  \caption{An example ARC-AGI task, ID: 00d62c1b. On the top are the input grids, with the bottom grids being their corresponding outputs. The task is to fill the insides of enclosed structures with yellow pixels.}
  \label{fg:arc} %
\end{figure}

The Abstraction and Reasoning Corpus (ARC; Figure~\ref{fg:arc}) \citep{chollet2019measure} offers one way to explore this. ARC is a set of 2D grid puzzles designed to be easy for humans but challenging for AI systems. Each ARC task is unique and only provides a few input-output examples from which the agent needs to learn to produce the correct output for the test grids. Deep learning methods on which the frontier AI models today are built rely on large amounts of data to learn representations without overfitting and struggle to generalize to out-of-distribution scenarios, making ARC particularly challenging due to the need for efficient and robust skill acquisition, which suggests the need for new learning paradigms. \footnote{On March 24, 2025, an updated benchmark, ARC-AGI-2, was released for the ARC Prize 2025 competition, which removed some tasks deemed susceptible to brute force search and includes new tasks designed to be more challenging for current AI reasoning systems. The evaluation and analysis in this paper are based on the original benchmark, used between 2020 and 2024.}

This motivates the search for alternatives that do not rely on scale or extensive priors. One such potential paradigm is a self-organizing system such as cellular automata. A cellular automaton (CA) is a discrete model of computation consisting of cells, typically organized in a grid, that are updated according to fixed rules that determine the new state of each cell based on the current states of the cell and its local neighbors.
Neural Cellular Automata (NCA) are an extension of traditional cellular automata, using a neural network in place of the discrete update rules. Inspired by the development of biological organisms from the self-organization of locally communicating cells, NCAs have been successfully used to perform the task of morphogenesis, growing and maintaining a target image, on a grid of pixels and have shown that robust regenerative capabilities can emerge from the constraints of self-organization \citep{mordvintsev2020growing}.

The appeal of NCAs lies in their simplicity and the constraints they impose. They perform computation through local, repeated updates, without any global coordination or structured memory. This setup naturally enforces certain inductive biases, such as translation invariance through shared convolutional kernels and natural support for spatial regularity. These same constraints also make NCAs difficult to train and sensitive to architecture choices. However, when they succeed, they often do so in ways that are compact, interpretable, and unexpectedly robust, properties that are often missing in conventional neural models.

Motivated by this robustness, this work explores the question at the beginning of this section by applying Neural Cellular Automata to tasks from ARC, introducing an exploratory approach to ARC, using NCAs to learn iterative update rules that transform input grids into their outputs from the training examples and apply them to the test inputs. The goal is not to solve ARC, but to use it as a diverse, high-pressure domain to study the capabilities and limits of NCAs as a model of computation. These tasks give a useful spectrum: some align naturally with NCA behavior, others reveal structural or functional bottlenecks that help clarify what this architecture can and cannot do.

Rather than aim at state-of-the-art performance, the focus is on analyzing the kinds of behavior these models can learn, how they generalize or fail to, and what this reveals about their strengths and limitations, some specific to this domain and others identifying broader challenges in NCA applications. In doing so, the intent is to surface deeper questions about architecture, spatial structure, and local interaction, and outline promising directions for future research into NCA's potential as a learning model.

\section{Related Work}

The ideas behind this work intersect with multiple lines of research, between self-organizing systems such as NCAs, and approaches to the ARC challenge. While the primary focus is on understanding how decentralized, locally-updated models behave in abstract transformation tasks, it is important to position this work within the broader landscape of related methods, motivations, and prior research. The results in this paper are not meant to be conclusive, but rather to surface insights that may inform future directions. Other applications of NCAs or self-organization in general, though technically different, often rely on similar techniques and may intersect meaningfully with the observations made from this work.

\subsection{Self-Organizing Systems}
Cellular automata have been extensively studied for decades by researchers observing complex behaviors emerging from simple rules. Notably, Conway's Game of Life \citep{gardner1970life} and Wolfram's Rule 110 \citep{wolfram2002nks} are two cellular automata that are known to be Turing complete, demonstrating the potential for these models to perform complex tasks \citep{cook2004universality}. \citet{crutchfield2002collective} studied using a genetic algorithm to evolve CA rules that perform density classification, a global computational task.
``Creatures" discovered in Lenia, a continuous generalization of Conway's Game of Life, were shown to be capable of expressing lifelike survival behaviors \citep{chan2019lenia}.

Researchers have also observed the connection between cellular automata and convolutional neural networks, using convolutions to represent the local interactions of cells \citep{gilpin2019cacnn}.  Neural Cellular Automata represent a more recent advancement in this field, enabling the differentiable learning of update rules to incorporate ideas from modern deep learning.
While initially introduced as a self-organizing model of morphogenesis, NCAs have also been studied under a variety of other tasks such as MNIST digit classification, maze solving, image repair, and simple cart-pole control \citep{randazzo2020self-classifying, endo2021maze, variengien2021selforganizedcontrolusingneural}. This paper extends these results by investigating how NCAs perform when applied to tasks requiring precise transformations and few-shot generalization, using the ARC benchmark as a domain that challenges their capabilities in ways not previously explored.

\subsection{ARC Approaches}
Designed to be challenging for AI, the strongest early approaches to the ARC tasks were based on program search, typically using a Domain-Specific Language (DSL) constructed specifically for ARC. However, this is an exponentially large search space, so deep-learning-guided search is the popular area of focus. The most successful approaches from the 2024 ARC-Prize competition were based on large language models pioneering test-time adaptation, with little progress prior to this. While plenty of other interesting directions exist, there is less focus on them---particularly on approaches that emphasize computational efficiency or do not make use of extensive pretraining of ARC-specific data. The Neural Cellular Automata approach proposed in this paper is an example of moving away from the conventional, centralized computing paradigm of deep learning.

Cellular automata rules can also be viewed as a program that is applied repeatedly until some stop condition is met. As such, the evolutionary search for CA rules can be seen as a form of program search. Early genetic programming work explored this idea for finding discrete CA rule sets for tasks such as the majority problem (density classification). %
Similarly, a Neural Cellular Automata approach to ARC could be seen as a differentiable program search.

One other work, CAX, experimented with training NCAs on 1D-ARC \citep{faldor2024cax, xu2024llmsabstractionreasoningcorpus}, and outperformed GPT-4 on it. 1D-ARC is an adaptation of the ARC benchmark to simplified one-dimensional tasks rather than 2D grids. While it does give some insight on what NCAs struggle to learn, such as counting tasks, these tasks are significantly less complex and do not represent the combination of multiple complex priors the way ARC tasks do. The original ARC benchmark is therefore used as the domain of interest in this paper.

Concurrent to this work, \citet{guichard:alife25} independently applied Neural Cellular Automata to ARC, introducing an augmented architecture with hidden memory states. Their focus was on leveraging this extension to enhance the performance of standard NCAs, whereas this work seeks to understand the capabilities and limits of NCAs by examining their characteristics on ARC tasks, and establish directions for future work in self-organizing computation.

\section{Approach}

The goal of this work is not to propose a finalized architecture for ARC, but to study how a self-organizing system like an NCA behaves when tasked with ARC's abstract grid transformations. What follows is a natural implementation of NCAs adapted to ARC formats, with minimal hand-engineering. Many variations in architecture and task setup are possible, but the focus is on a straightforward implementation that serves as a natural baseline. This approach makes it possible to examine what kinds of structure, generalization, and failure arise from the model itself, without relying on external priors or domain-specific tuning. The approach draws from previous NCA work where applicable, making it easier to compare observations across domains and to carry insights in two directions: adapting known techniques to ARC, and identifying properties that may inform future NCA applications more broadly. Where relevant, variants that were tested but were found to underperform or destabilize training are noted.

\subsection{Architecture}

The architecture used is similar to the one used by \citet{mordvintsev2020growing}, but with a few key modifications to adapt it to the ARC domain. ARC grids are given as 2D arrays of integers 0-9, representing one of 10 possible colors per cell. This is converted to a one-hot encoding for the initial state of the NCA, such that for a cell at any given time step, the \texttt{argmax} of the first 10 channels represents the color being expressed (in practice, a \texttt{softmax} is used for differentiability).
Following this categorical structure, rather than an incremental update of the form
\[
s_{t+1} = s_t + f_\theta(s_t) \text{,} %
\]
a new state is output by the network each step as
\[
s_{t+1} = f_\theta(s_t) \text{.}
\]

The remaining channels are simply hidden states that can represent any kind of information that is needed and are not passed through a \texttt{softmax} function. The number of hidden channels is a hyperparameter, with more channels allowing for increased complexity, but a larger number of parameters to be optimized.

Additionally, a standard convolutional layer with learnable parameters is used in place of the fixed, depthwise Sobel filters to provide more flexibility. In addition, \texttt{layernorms}, i.e. normalization of activations across channels, are used to help stabilize training.

\subsection{Asynchronous Update Scheme}

Stochastic cell updates are used similarly to the original implementation \citep{mordvintsev2020growing} so that global synchronization is not necessary. By randomly selecting a portion of cells to update at each step, asynchronicity is introduced to the system. This makes the task more difficult, as correct updates must be robust to surrounding cells being out of sync, but is known to provide a regularization that improves robustness to noise and damage for morphogenesis \citep{mordvintsev2020growing, miotti2025difflogic}. Given sufficient additional time steps during training to handle the asynchronicity, an NCA is usually able to learn the same end behavior with a slight decrease in consistency. In the experiments in this paper, this constraint is relaxed even further: after selecting the cells to have their updates masked by some predefined probability, rather than fully setting their update values to zero, the new state is interpolated with the previous state by a random strength between 0 and 1. This change maintains the stochastic asynchronicity but makes it smoother when the total number of time steps is small.

It is rather surprising that the model is still able to learn under highly asynchronous conditions, and in practice many solved tasks do not seem to require stochastic updates, but overall accuracy and consistency are improved across the benchmark. Thus, stochastic cell updates act as a form of regularization by filtering 'weak' solutions that overfit to the training examples without learning anything resembling the correct, generalizable transformation. This effect also suggests that other architectural constraints that similarly help identify stronger solutions, possibly with domain-specific biases, may be promising future improvements. A related idea from NCAs for morphogenesis \citep{mordvintsev2020growing} is to keep track of whether each cell is alive or not: only a single seed cell starts as alive, and dead cells are not updated. This mechanism was not found to provide any improvement in preliminary experiments and was not used in the experiments below.

\subsection{Training}
The input for a task is treated as the initial state of the NCA. A pixel-wise log loss is used to compare the first 10 channels of each cell with the one-hot encoding of the expected output grid. Training is done with standard backpropagation-through-time over a fixed number of time steps, across all given training examples. Following this recurrent training analogy, it may seem natural to evaluate the loss function only at the final time step and not supervise intermediate transitions. However, loss is applied at every step for three reasons that proved important in practice:

\begin{enumerate}

\item For simpler tasks that do not rely on long distance organization, such as those that are solvable within one time-step, taking loss only on the final time step does not encourage the NCA to converge and stabilize as early as possible. As a result, the intermediate states often become corrupted and have to recover, which makes generalization harder. This kind of failure is observed in as little as 15 time steps, significantly fewer than the number of steps used for morphogenesis. Meanwhile, other tasks may benefit from or require a higher number of time steps, and thus this issue is not acceptable for an approach that is generally applicable.

\item When training on a higher number of time steps, measuring loss only on the final one can lead to a problem with gradient flow. One solution to this is to task the network to perform incremental, additive updates, as in \citet{mordvintsev2020growing}, rather than requiring it to output a new state at every step. However, this modification still turns out to underperform the current approach in stability and generalization with no clear benefit.

\item Small errors tend to compound over time; constant supervision mitigates this problem. Hidden channels are also still allowed to change freely, preventing the first 10 channels from having to simultaneously handle both information propagation and loss minimization, which could otherwise lead to unstable or entangled behavior. However, such training still makes convergence greedy, and the approach could miss solutions that require more complex dynamics.

\end{enumerate}

In addition to constant supervision, varying the mask probability of stochastic cell updates during training proved useful in preliminary experiments as well. Instead of using a fixed probability such as 0.5, as commonly done in prior work \citep[e.g.][]{mordvintsev2020growing}, each training rollout samples a value from a specified range (typically between 0 and 0.75). Models trained with a fixed mask probability often do not consistently converge. By exposing the model to a range of update rates during training, it learns to perform reliably under different conditions. As a result, the same model can be run with fully synchronous updates (i.e., no masking) at test time and still produce stable outputs, improving evaluation consistency that would otherwise be undermined by stochastic asynchronicity.

Notably, even on failed tasks, models trained with varied mask probabilities often produced more coherent or partially correct outputs, suggesting a form of improved generalization despite not fully solving the task.
An additional benefit of this strategy is improved long-term stability, similar in effect to the persistence techniques used in single-target morphogenesis tasks. In prior work, \cite{mordvintsev2020growing} used a sample-pool training method, where intermediate outputs are reused as new starting points to help the model refine and stabilize its behavior over time. In contrast, varying the mask probability offers a simpler and broadly applicable way to expose the model to a range of intermediate dynamics.

Although ARC evaluations in this paper typically use only a small number of update steps, and no persistence training techniques were applied, the stability benefits observed from varying the mask probability suggest a broader role for this strategy. Persistence-based methods, such as sample-pool training in morphogenesis, explicitly train models to recover from incomplete or perturbed states by reinforcing convergence toward a fixed target. While the approach used for ARC achieves stability through exposure to varied update schedules rather than explicit persistence, both methods seem to promote the same underlying property: robust convergence from a broad range of noisy or intermediate states. Learning to recover from these imperfect conditions may help the model develop a transformation function that closer represents the task's intended rule and generalizes better. The observed connection between long-term stability and improved generalization points to a useful insight for future ARC approaches and other applications of self-organizing systems.

As seen in this paper and noted in other works \citep{miotti2025difflogic}, training NCAs to produce complex patterns is challenging and often requires extensive hyper-parameter tuning. One goal of this paper is to identify possible sources of this difficulty and potential directions to address them.

\section{Experiments}

This section presents the results of training a model for each task from the public training set of ARC. It should be noted that each task is unique and they vary heavily, so benchmark score (number of passed tasks) is not a sufficient metric to characterize and examine NCA behavior. Additionally, minor changes to hyperparameters can strongly affect the pass/fail consistency of a task.

\subsection{Task Restrictions}

Note that the current approach includes no mechanism for growing or shrinking the grid. Thus, it is restricted to tasks where each input-output pair has matching grid dimensions, though the grid size itself may vary across examples.

Also, some ARC tasks require color generalization, where a color is used in test inputs but does not appear in training. Because colors are distinct channels in the network, such tasks are almost always guaranteed to fail. A discrete Cellular Automata rule based solution might consist of rules that apply to all colors, i.e., multiple states, but this behavior is less natural for an NCA that simply outputs 10 states and learns from gradient descent.

However, these limitations do not inherently restrict the types of transformations of interest as they relate to generalization and behavior, so even without being able to examine these affected tasks, it is likely that the insights from the remaining tasks are still applicable. %

\subsection{Setup}

The same hyperparameter settings were used for all tasks. They were found experimentally to work well and be consistent, but not optimized separately for each task. They were: 20 hidden channels, 10 time steps allowed, stochastic update mask probability range of [0, 0.75], and 128 trials for each example per training epoch.

The models were trained for 800 epochs using the AdamW \citep{loshchilov2019decoupledweightdecayregularization} optimizer, with a learning rate linearly decayed from 0.002 to 0.0001. They have just over 10,000 parameters. In practice, much fewer than 20 hidden channels were needed for a majority of tasks, allowing models smaller than 2500 parameters to perform similarly in some cases.

\subsection{Overall Performance}

Of the 400 public training tasks, 138 were out of scope because of the input-output grid size restriction. Another 90 tasks involved colors in their test inputs that do not appear in training grids, and would all fail without revealing anything about NCA behavior. Thus, there were 172 feasible tasks. Of these, the NCA approach solved 23 perfectly, resulting in a success rate of 13.37\%.

While this number reflects the overall performance of the approach, it offers little insight into how many tasks were nearly solved or how many exhibit high variance and require parameter tuning to succeed consistently.
Moreover, meaningful comparisons to other ARC approaches are difficult from this number alone. These experiments involve no pretraining and are run with minimal computational overhead, training each task-specific model from scratch in just a few minutes. Tasks are viewed as independent rather than as part of a training set for learning ARC's priors, so the goals and insights are fundamentally distinct from methods relying on large-scale search or pretraining.

Indeed, the aim is not to compete with such systems, but to study a fundamentally different approach to problem-solving through local, self-organizing computation. Therefore, a closer examination of tasks will be more insightful than overall performance, as will be illustrated in the next subsection. Finally, on 95 of the attempted tasks, training achieves a categorical cross entropy loss $< 0.01$, with only half of the successful tasks in this group. An additional 48 tasks had over 90\% of pixels in the end grid correct. While these metrics are not perfect signals of success, they further underscore the value of deeper analysis into the behaviors and limitations of this approach.

\subsection{Case Studies}
The range of task types are what makes ARC interesting for studying self-organization. The cases below were chosen to provide tangible insight into NCA's traits, successes, and failures.

\subsubsection{Spiral Generalization}

\begin{figure}[t]
  \centering
  \includegraphics[width=\linewidth]{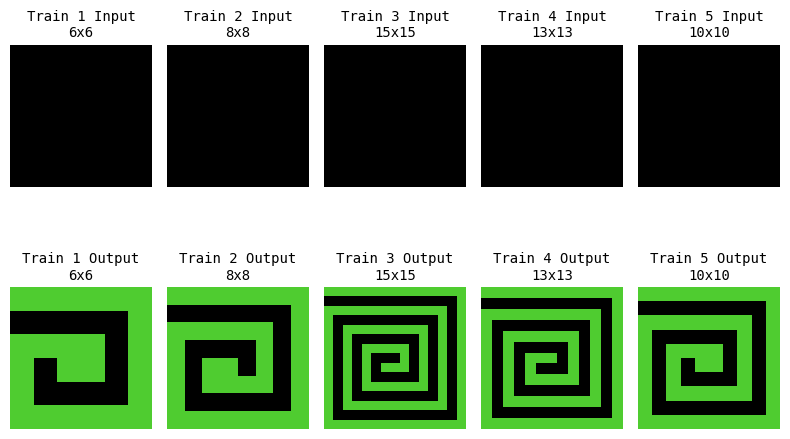}
  \caption{Task ID: 28e73c20. The input-output training pairs of the spiral task, used to learn the iterative update rule to construct any sized spiral.}
  \label{fg:spiral} %
\end{figure}

The spiral drawing task (\Cref{fg:spiral}) is visually interesting and helps illustrate what a self-organizing solution looks like. It is simple to test the quality of the learned cellular automata by scaling up the input grid, thus evaluating the generality and precision of the local interactions learned from smaller spirals.

\begin{figure*}[t]
  \centering
  \includegraphics[width=0.31\linewidth]{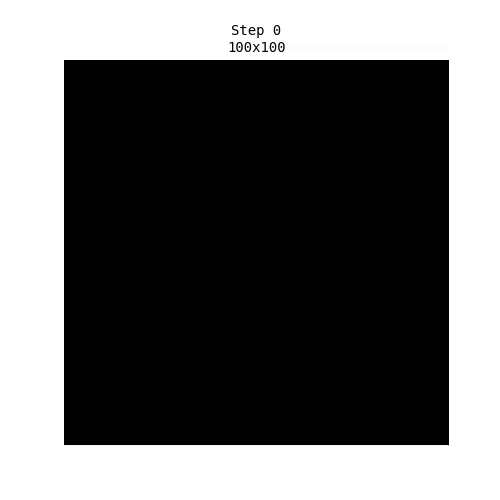}
  \hfill
  \includegraphics[width=0.31\linewidth]{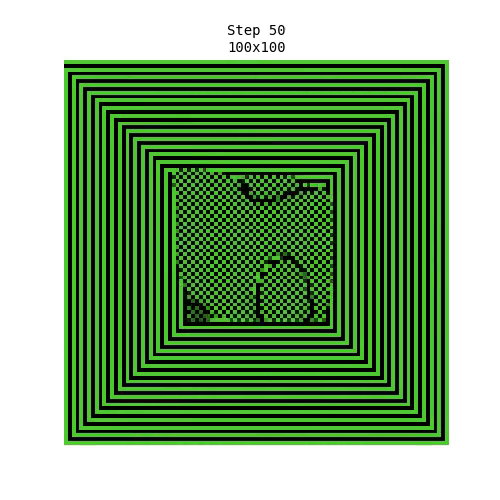}
  \hfill
  \includegraphics[width=0.31\linewidth]{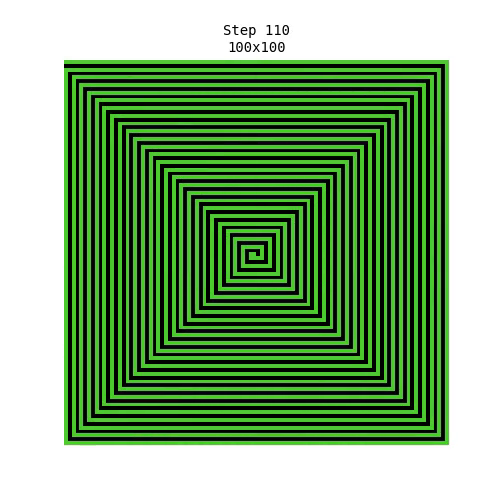}
  \caption{Spiral pattern generalized to 100$\times$100 grid, shown at three stages of the update process (t=0, 50, 110). The model was trained only on the small examples from \Cref{fg:spiral} and tested \textit{with} asynchronous updates on, yet maintains structural coherence across space and time, indicating strong generalization and stability.}
  \label{fg:spiral2}
\end{figure*}

Synthesizing a program to output a spiral given these examples would not be trivial, yet this task is surprisingly simple for cellular automata. The system learned a compact, iterative rule that is precise enough to have an effective degree of size invariance, capable of scaling up to a 100$\times$100 grid (\Cref{fg:spiral2}). Encoding the solution as local state transitions demonstrates how complexity of a transformation in a NCA differs fundamentally from that in centralized models. A possible explanation for the success on this task is that it is one with a consistent initial state and relatively few transition states that need to be learned. The rules are applied iteratively and there is little structural variation or noise that lead to overfitting. The only difference between inputs is grid size, which mostly only influences how many transition steps are needed before the full grid is settled. 
It is not always obvious what structures or operations are more naturally and efficiently represented by the properties of a particular self-organizing system, making them an interesting direction for benchmarks like ARC and broader computational modeling.

\subsubsection{Stochastic Update Regularization}

As stated previously, many of the successfully solved tasks do not rely on asynchronous updates. However, some tasks demonstrate that the regularization provided by asynchronicity can make a difference.

\begin{figure}[t]
  \centering
  \includegraphics[width=\linewidth]{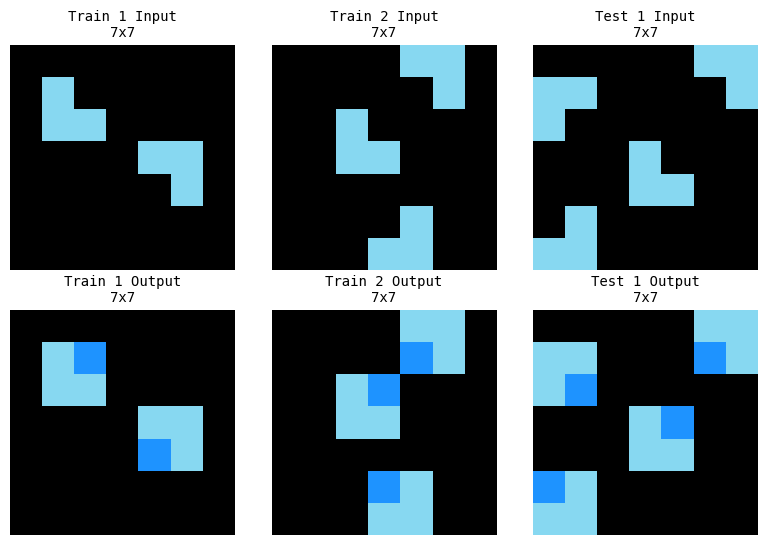}
  \caption{Task ID: 3aa6fb7a. Input-output pairs. The left two are training pairs, and the right pair is the test task. This task is structurally simple, filling in the corner of 2$\times$2 squares, but one of four orientations does not appear in the training examples, which is a problem for Neural Cellular Automata.}
  \label{fg:sur} %
\end{figure}

For instance, \Cref{fg:sur} demonstrates a very simple task that only requires a single update step. However, a simple NCA will fail to learn given the training examples. With the regularization provided by stochastic cell updates, the model, surprisingly, is able to handle this case, i.e.\ generalize to inputs that did not appear in training.

\subsubsection{Failure Cases}
Most of the failed tasks can be put into one of three categories: (1) models solve new examples inconsistently with high variance or only under specific settings, (2) models solve the training examples but fail to generalize to any new examples, and (3) models fail to even memorize the training examples. 

The second category is the largest, and indicates overfitting. With so few training examples per task, it is often unclear whether a correct general solution is even representable within the current architecture, or if the model is simply overfitting to the training cases. In many instances, failure could reflect either a genuine architectural limitation or just insufficient supervision to guide learning. With more examples, some of these tasks might converge on the correct transformation rather than memorizing input-output pairs.

\begin{figure}[t]
  \centering
  \includegraphics[width=\linewidth]{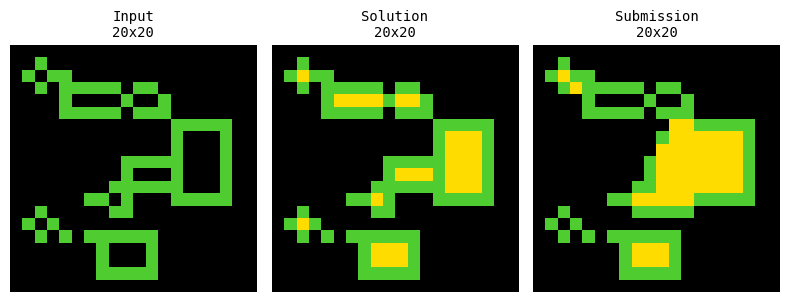}
  \caption{Task ID: 00d62c1b. The test input for the task in \Cref{fg:arc} is on the left, correct solution in the middle, and a trained model's output on the right. This case demonstrates a failure to learn the correct transformation and structural interpretation.}
  \label{fg:failure} %
\end{figure}

\Cref{fg:failure} is an example of a task where an NCA can successfully reproduce the training outputs, but does not correctly solve the test inputs. The model does not learn the correct transformation, but instead appears to memorize the training inputs, which becomes apparent when a new input is substantially structurally different. Even when it is possible for the correct transformation to be encoded by an NCA, supervised training with this setup will settle for any solution that fits the training examples. To generalize, the constraints imposed during training need to be strong enough to bias toward the intended rule.

These observations are discussed further in the next section, leading to ideas how the issues could be mitigated.

\section{Discussion and Future Work}

This paper explored Neural Cellular Automata as a substrate for abstract grid-based tasks from the ARC benchmark. While overall task coverage was limited, the selected examples revealed diverse behaviors. Certain tasks showed clear evidence of generalization beyond training, while others exposed brittleness or structural overfitting. These findings clarify the strengths and weaknesses of NCAs for spatial reasoning tasks, and suggest promising paths forward.

\subsection{Stability and Generalization}

A recurring observation in this study was the benefit of training NCAs under varying conditions to improve stability and generalization. Training models using varying stochastic mask probabilities improved their robustness, enabling convergence from noisy or unexpected intermediate states. Even in cases that ultimately failed to pass the full test task, models trained this way produced outputs that were more coherent and closer to correct.

This stability is similar in principle to persistence-based methods like the sample-pool approach used in NCA morphogenesis. Both methods encourage models to develop stable attractors rather than memorizing specific update sequences. The intuition here is straightforward: by training models to recover from imperfect conditions, they are more likely to learn generalizable rules. In other words, introducing noise or variation in training helps filter out fragile solutions, favoring robust and general solutions that reflect more meaningful internal states.

\subsection{Failure Modes and Limitations}

Despite promising outcomes, some clear limitations of the current approach also emerge. First, NCAs struggled significantly with tasks that required global coordination or propagation of information across longer distances. Because NCAs rely purely on local interactions, cells frequently updated prematurely before global context was adequately established, causing structural overfitting.

Perhaps more fundamentally, the approach described here faces a challenge analogous to one seen in program synthesis: there are often many possible solutions that can fit the provided training examples, but most solutions are not the correct transformation agreed on by most humans and will fail to generalize. In classical program search, one common approach is to prefer the simplest program that fits the data, using a heuristic like Kolmogorov complexity. However, within the gradient-based training paradigm of NCAs, there is no straightforward way to measure or control for this kind of ``solution complexity." Because the architecture and parameter space are fixed and continuous, gradient descent will settle on any plausible local optimum, often based on random initialization, rather than explicitly selecting the simplest or most general solution.

In fact, given the limited number of ARC training examples, memorizing individual examples might sometimes represent a simpler solution (under this encoding) than actually learning the correct transformation rule. This limitation implies that purely gradient-based learning of NCAs could be fundamentally missing something important about the problem structure. One potential way to address this is to explicitly minimize complexity, by attempting to find the smallest network that can solve the task. Another possible direction is to reframe the learning process in probabilistic terms. While speculative, incorporating ideas from Bayesian or variational inference could help. Instead of converging on a single solution, the model would learn a distribution over plausible behaviors and could favor simpler or more likely rules under uncertainty. This might provide a way to steer learning toward more generalizable solutions in low-data settings.

\subsection{Design Variations}

Several modifications and alternative designs were tested with limited success. For instance, incremental hidden-state updates performed worse in practice compared to directly outputting the new state each step, possibly due to saturation issues or vanishing gradients. Similarly, skip connections and alternative activation functions were inconsistent or provided minimal improvements.

The choice of number of time steps is another factor with noticeable impact. For benchmarking, 10 steps were used for consistency, though this setting is somewhat arbitrary and not optimal for every task. Some tasks, like the spiral, generalized more reliably to larger grids (e.g., 100$\times$100) when trained with 30 steps, despite identical training examples. This suggests that certain transformations benefit from longer iterative computation, and that fixed step counts may limit what the model can express.

\subsection{Efficiency and Comparison to Existing Approaches}

A key strength of this approach is its computational efficiency. Models trained in these experiments required only minutes per task on a single consumer GPU. This contrasts sharply with ARC approaches using large language models or other deep learning architectures, which typically involve extensive pretraining or many hours of computation.

However, direct comparison with these approaches is not straightforward or particularly meaningful due to fundamental differences in assumptions and constraints. Instead of seeing these approaches as competing ideas, it may be more productive to consider how their insights and techniques might integrate with NCAs and potentially address limitations of the current approach.

\subsection{Future Work}

Based on the insights from the results of this work, some promising directions for future work stand out:

\begin{itemize}
    \item \textbf{Adaptive computation strategies:} Dynamically adjusting computation time steps \citep{graves2017adaptivecomputationtimerecurrent} may help due to the observation that fixed step counts may be limiting.
    \item \textbf{Variational and probabilistic NCAs:} Leveraging Bayesian methods to explicitly manage complexity and select more generalizable solutions. Other neural approaches to ARC-AGI such as Latent Program Networks \citep{bonnet2024searchinglatentprogramspaces} and CompressARC \citep{liao2025arcagiwithoutpretraining} use these ideas, and variational NCAs have been studied in prior work \citep{palm2022variationalneuralcellularautomata}.
    \item \textbf{Interpretability:} Developing methods for evaluating and interpreting model behavior to better understand the limitations of this architecture. Discrete variations like Difflogic CA \citep{miotti2025difflogic} as well as more traditional forms of self-organization such as evolving CA rulesets are advantaged in this sense. 

    \item \textbf{Architectural improvements:} Exploring ideas from mainstream neural network research, such as attention, gating, or adaptive computation, that address similar weaknesses in recurrent models but have not yet been widely explored with self-organizing models. Given their structural similarities to other neural architectures, NCAs could benefit from adapting these techniques in ways that preserve local update dynamics.

    \item \textbf{Broader evaluations:} Testing beyond ARC to other discrete and continuous reasoning benchmarks to better characterize the generalization capabilities of NCAs.
\end{itemize}

\subsection{Broader Implications}

Certain tasks from ARC highlighted NCA's strong potential for robust, generalizable learning from local interactions alone. Further improving NCAs in terms of stability, flexibility, and interpretability can benefit both ARC-specific research and the broader study of decentralized, self-organizing computation. Working towards these improvements could lead to general strategies for developing robust, scalable, and adaptive neural systems beyond the context of ARC.

\footnotesize
\bibliographystyle{apalike}
\bibliography{nca_arc_paper} %

\begin{thebibliography}{}

\bibitem[Bonnet and Macfarlane, 2024]{bonnet2024searchinglatentprogramspaces}
Bonnet, C. and Macfarlane, M.~V. (2024).
\newblock Searching latent program spaces.
\newblock {\em arXiv:2411.08706}.

\bibitem[Chan, 2019]{chan2019lenia}
Chan, B. W.-C. (2019).
\newblock Lenia—biology of artificial life.
\newblock {\em Complex Systems}, 28(3):251--286.

\bibitem[Chollet, 2019]{chollet2019measure}
Chollet, F. (2019).
\newblock On the measure of intelligence.
\newblock {\em arXiv:1911.01547}.

\bibitem[Cook, 2004]{cook2004universality}
Cook, M. (2004).
\newblock Universality in elementary cellular automata.
\newblock {\em Complex Systems}, 15(1):1--40.

\bibitem[Crutchfield et~al., 2002]{crutchfield2002collective}
Crutchfield, J.~P., Mitchell, M., and Das, R. (2002).
\newblock The evolutionary design of collective computation in cellular automata.
\newblock In Crutchfield, J.~P. and Schuster, P.~K., editors, {\em Evolutionary Dynamics—Exploring the Interplay of Selection, Neutrality, Accident, and Function}, pages 361--411. Oxford University Press, Oxford, UK.

\bibitem[Endo and Yasuoka, 2021]{endo2021maze}
Endo, K. and Yasuoka, K. (2021).
\newblock Neural cellular maze solver.
\newblock \url{https://umu1729.github.io/pages-neural-cellular-maze-solver/}.
\newblock Retrieved 2025-08-15.

\bibitem[Faldor and Cully, 2024]{faldor2024cax}
Faldor, M. and Cully, A. (2024).
\newblock {CAX:} {C}ellular automata accelerated in \textsc{JAX}.
\newblock {\em arXiv:2410.02651}.

\bibitem[Gardner, 1970]{gardner1970life}
Gardner, M. (1970).
\newblock {Mathematical Games}: The fantastic combinations of {John Conway}’s new solitaire game ``{Life}''.
\newblock {\em Scientific American}, 223(4):120--123.

\bibitem[Gilpin, 2019]{gilpin2019cacnn}
Gilpin, W. (2019).
\newblock Cellular automata as convolutional neural networks.
\newblock {\em Physical Review E}, 100(3):032402.

\bibitem[Graves, 2017]{graves2017adaptivecomputationtimerecurrent}
Graves, A. (2017).
\newblock Adaptive computation time for recurrent neural networks.
\newblock {\em arXiv:1603.08983}.

\bibitem[Guichard et~al., 2025]{guichard:alife25}
Guichard, E., Reimers, F., Kvalsund, M., Lepperød, M., and Nichele, S. (2025).
\newblock {ARC-NCA}: Towards developmental solutions to the abstraction and reasoning corpus.
\newblock In {\em Proceedings of the 2025 Conference on Artificial Life (ALIFE 2025)}.

\bibitem[Liao and Gu, 2025]{liao2025arcagiwithoutpretraining}
Liao, I. and Gu, A. (2025).
\newblock {ARC-AGI} without pretraining.
\newblock \url{https://iliao2345.github.io/blog_posts/arc_agi_without_pretraining/arc_agi_without_pretraining.html}.
\newblock Retrieved 2025-08-15.

\bibitem[Loshchilov and Hutter, 2019]{loshchilov2019decoupledweightdecayregularization}
Loshchilov, I. and Hutter, F. (2019).
\newblock Decoupled weight decay regularization.
\newblock In {\em International Conference on Learning Representations}.

\bibitem[Miotti et~al., 2025]{miotti2025difflogic}
Miotti, P., Niklasson, E., Randazzo, E., and Mordvintsev, A. (2025).
\newblock Differentiable logic cellular automata: From game of life to pattern generation.
\newblock {\em arXiv:2506.04912}.

\bibitem[Mordvintsev et~al., 2020]{mordvintsev2020growing}
Mordvintsev, A., Randazzo, E., Niklasson, E., and Levin, M. (2020).
\newblock Growing neural cellular automata.
\newblock {\em Distill}.

\bibitem[Palm et~al., 2022]{palm2022variationalneuralcellularautomata}
Palm, R.~B., González-Duque, M., Sudhakaran, S., and Risi, S. (2022).
\newblock Variational neural cellular automata.
\newblock {\em arXiv:2201.12360}.

\bibitem[Randazzo et~al., 2020]{randazzo2020self-classifying}
Randazzo, E., Mordvintsev, A., Niklasson, E., Levin, M., and Greydanus, S. (2020).
\newblock Self-classifying {MNIST} digits.
\newblock {\em Distill}.
\newblock https://distill.pub/2020/selforg/mnist.

\bibitem[Variengien et~al., 2021]{variengien2021selforganizedcontrolusingneural}
Variengien, A., Nichele, S., Glover, T., and Pontes-Filho, S. (2021).
\newblock Towards self-organized control: Using neural cellular automata to robustly control a cart-pole agent.
\newblock {\em arXiv:2106.15240}.

\bibitem[Wolfram, 2002]{wolfram2002nks}
Wolfram, S. (2002).
\newblock {\em A New Kind of Science}.
\newblock Wolfram Media, Champaign, IL.

\bibitem[Xu et~al., 2024]{xu2024llmsabstractionreasoningcorpus}
Xu, Y., Li, W., Vaezipoor, P., Sanner, S., and Khalil, E.~B. (2024).
\newblock Llms and the abstraction and reasoning corpus: Successes, failures, and the importance of object-based representations.
\newblock {\em arXiv:2305.18354}.

\end{thebibliography}

\end{document}